**BMC Bioinformatics**

**RESEARCH** **Open Access**

# CollaboNet: collaboration of deep neural networks for biomedical named entity recognition

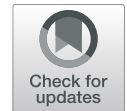

Wonjin Yoon[1†], Chan Ho So[2†], Jinhyuk Lee[1] and Jaewoo Kang[1,2*]

## Abstract

**Background:** Finding biomedical named entities is one of the most essential tasks in biomedical text mining. Recently, deep learning-based approaches have been applied to biomedical named entity recognition (BioNER) and showed promising results. However, as deep learning approaches need an abundant amount of training data, a lack of data can hinder performance. BioNER datasets are scarce resources and each dataset covers only a small subset of entity types. Furthermore, many bio entities are polysemous, which is one of the major obstacles in named entity recognition.

**Results:** To address the lack of data and the entity type misclassification problem, we propose CollaboNet which utilizes a combination of multiple NER models. In CollaboNet, models trained on a different dataset are connected to each other so that a target model obtains information from other collaborator models to reduce false positives. Every model is an expert on their target entity type and takes turns serving as a target and a collaborator model during training time. The experimental results show that CollaboNet can be used to greatly reduce the number of false positives and misclassified entities including polysemous words. CollaboNet achieved state-of-the-art performance in terms of precision, recall and F1 score.

**Conclusions:** We demonstrated the benefits of combining multiple models for BioNER. Our model has successfully reduced the number of misclassified entities and improved the performance by leveraging multiple datasets annotated for different entity types. Given the state-of-the-art performance of our model, we believe that CollaboNet can improve the accuracy of downstream biomedical text mining applications such as bio-entity relation extraction.

**Keywords:** NER, Deep learning, Named entity recognition, Text mining

## Background

The amount of biomedical text continues to increase rapidly. There were 4.7 million full-text online accessible articles in PubMed Central [1] in 2017. One of the obstacles in utilizing biomedical text data is that it is too large for a human to read or even search for needed information. This has led to the demand for automated extraction of valuable information. Text mining can be used to turn the time-consuming task into a fully automated job [2–7].

Named Entity Recognition (NER) is the computerized procedure of recognizing and labeling entities in given texts. In the biomedical domain, typical entity types include disease, chemical, gene and protein.

Biomedical named entity recognition (BioNER) is an essential building block of many downstream text mining applications such as extracting drug-drug interactions [8] and disease-treatment relations [9]. BioNER is also used when building a sophisticated biomedical entity search tool [10] that enables users to pose complex queries to search for bio-entities.

*Correspondence: kangj@korea.ac.kr
†Wonjin Yoon and Chan Ho So contributed equally to this work.
[1]Department of Computer Science and Engineering, Korea University, 02841 Seoul, Republic of Korea
[2]Interdisciplinary Graduate Program in Bioinformatics, Korea University, 02841 Seoul, Republic of Korea

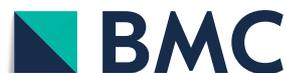

NER in biomedical text mining is focused mainly on dictionary-, rule-, and machined learning-based approaches [11–16]. Dictionary based systems have a simple and intuitive structure but they cannot handle unseen entities or polysemous words, resulting in low recall [11, 12]. Moreover, building and maintaining a comprehensive and up-to-date dictionary involves a considerable amount of manual work. The rule based approach is more scalable, but it needs hand crafted feature sets to fit a model to a dataset [13, 14]. These rule and dictionary-based approaches can achieve high precision [10] but can produce incorrect predictions when a new word, which is not in the training data, appears in a sentence (out-of-vocabulary problem). This out-of-vocabulary problem occurs frequently especially in the biomedical domain, as it is common for a new biomedical term, such as a new drug name, to be registered in this domain.

Recently, studies have demonstrated the effectiveness of deep learning based methods. Sahu and Anand [17] demonstrated the efficiency of Recurrent Neural Network (RNN) for NER in biomedical text. The model by Sahu and Anand is composed of a bidirectional Long Short-Term Memory Network (BiLSTM) and Conditional Random Field (CRF). Sahu and Anand [17] also used character level word embeddings but could not demonstrate their benefits. Habibi et al. [18] combined the BiLSTM-CRF model implementation of Lample et al. [19] and the word embeddings of Pyysalo et al. [20]. Habibi et al. [18] utilized character level word embeddings to capture characteristics, such as orthographic features, of bio-medical entities and achieved state-of-the-art performance, demonstrating the effectiveness of character level word embeddings in BioNER.

Although these models showed some promising results, NER is still a very challenging task in the biomedical domain for the following reasons. First, a limited amount of training data is available for BioNER tasks. Gold-standard datasets contain annotations of one or two entity types. For example, the NCBI corpus [21] includes annotations of diseases but not of other types of entities such as genes and proteins. On the other hand, the JNLPBA corpus [22] contains annotations of only genes and proteins. Therefore, the data for each entity type comprises only a small portion of the total amount of annotated data.

Multi-task learning (MTL) is a method for training a single model for multiple tasks at the same time. MTL can leverage different datasets that are collected for different but related tasks [23]. Although extracting genes is different from extracting chemicals, both tasks require learning some common features that can help understand the linguistic expressions of biomedical texts. Crichton et al. [24] developed an MTL model that was trained on various source datasets containing annotations of different subsets of entity types. An MTL model by Wang et al. [25] achieved performance comparable to that of the state-of-the-art single task NER models. Inspired by the previous studies, we propose CollaboNet which uses the collaboration of multiple models. Unlike the conventional MTL methods which use only a single static model, CollaboNet is composed of multiple models trained on different datasets for different tasks. Each model in CollaboNet is trained on dataset annotated on a specific type of entity and becomes an expert on their own entity type.

Despite the high recall obtained by the MTL based models, the precision of these models is relatively low. Since MTL based models are trained on multiple types of entities and larger training data, they have a broader coverage of various biomedical entities, which naturally results in high recall. On the other hand, as the MTL models are trained on combinations of different entity types, they tend to have difficulty in differentiating among entity types, resulting in lower precision.

Another reason NER is difficult in the biomedical domain is that an entity could be labeled as different entity types depending on its textual context. In our experiments, we observed that many incorrect predictions were a result of the polysemy problem, in which a word, for example, can be used as both a gene and disease name. Models designed to predict disease entities misidentify some genes as diseases. This misidentification of entity types increases the false positive rate. For instance, BiLSTM-CRF based models for disease entities mistakenly label the gene name *"BRCA1"* as a disease entity because there are disease names such as *"BRCA1 abnormalities"* or *"Brca1-deficient"* in the training set. Besides, the training set that annotates *"VHL"* (Von Hippel-Lindau disease) as a disease entity confuses the models because VHL is also used as a gene name, since the mutation of this gene causes VHL disease.

To solve the false positive problems due to polysemous words, CollaboNet aggregates the results of collaborator models, and uses them as an additional input to the target model. Consider the case of predicting the disease entity VHL utilizing the outputs of gene and chemical models. Once a gene model predicts VHL as a gene, the gene model informs a disease model that VHL is a gene entity so that the disease model will not predict VHL as a disease. In CollaboNet, each model is individually trained on an entity type and then further trained on the outputs of other models that are trained on the other entity types. The models in CollaboNet take turns in being the target and collaborator models during training. Consequently, each model is an expert in its own domain and helps improve the accuracy by leveraging the multi-domain information from the other models.

## Methods

In the following section, we first discuss a BiLSTM-CRF model for biomedical named entity recognition. The overall structure of the BiLSTM-CRF model is illustrated in Fig. 1. Next, we introduce the structure of CollaboNet, which is comprised of a set of BiLSTM-CRF models as shown in Fig. 2.

### Problem Definition

Named entity recognition involves annotating words in a sentence as named entities. More formally, given an input sequence $S = [w_1, w_2, ..., w_N]$, we predict corresponding labels $Y = [y_1, y_2, ..., y_N]$. We use the BIOES scheme [26] for representing $y_t$, where B stands for Beginning, I for Inside, O for Out, E for End, and S for Single.

### Embedding layer

#### Word Embedding (WE)

Word embedding is an effective way of representing words. As word embeddings capture semantic and syntactic meanings of words, they have been widely used in various natural language processing tasks including named entity recognition. The experiment of Habibi et al. [18] showed that word embeddings trained on biomedical corpora notably improved the performance of BioNER models. Pyysalo et al. [20] were the first to suggest training word embeddings on biomedical corpora from PubMed, PubMed Central (PMC), and Wikipedia. The results of Pyysalo et al. [20] and Habibi et al. [18] suggest that using word embeddings trained on biomedical corpora is essential for BioNER. We also use the trained word embeddings provided by Pyysalo et al. [20]. For each word $w_t$ in a sequence $S$, we denote a word represented by a word embedding as $x_t \in \mathbb{R}^{d^{word}}$ where $d^{word}$ is a dimension of the word embedding.

#### Character Level Word Embedding (CLWE)

To give our model character level morphological information (*e.g.*, '*-ase*' is common in protein entities), we also leverage the character level information of each word. We build character level word embeddings (CLWEs) using a convolution neural network (CNN), similar to the work of Santos and Zadrozny [27]. Given a word $w_t$, composed of $M$ number of characters, we represent $w_t = \{c_1^t, c_2^t, ..., c_M^t\}$ where $c_i^t \in \mathbb{R}^{d^{char}}$ is a randomly initialized character embedding for each unique character. Note that unlike the word embeddings trained on separate biomedical corpora, character embeddings are learned from only the BioNER task. For the CNN, padding of the proper size $((k-1)/2)$ according to window size $k$ should be attached before and after each word. We obtain a window vector $C_i^t$ by simply concatenating the character embeddings of $c_i^t$ with the character embeddings of $(k-1)/2$ characters on both sides:

$$C_i^t = \left[c_{i-(k-1)/2}^t, \cdots c_i^t, \cdots c_{i+(k-1)/2}^t\right] \in \mathbb{R}^{kd^{char}} \quad (1)$$

From the window vector $C_i^t$, we perform a convolution operation as follows:

$$\left[x_t^c\right]_j = \max_{1 \le i \le M} \left[W_{char} C_i^t + b_{char}\right]_j \quad (2)$$

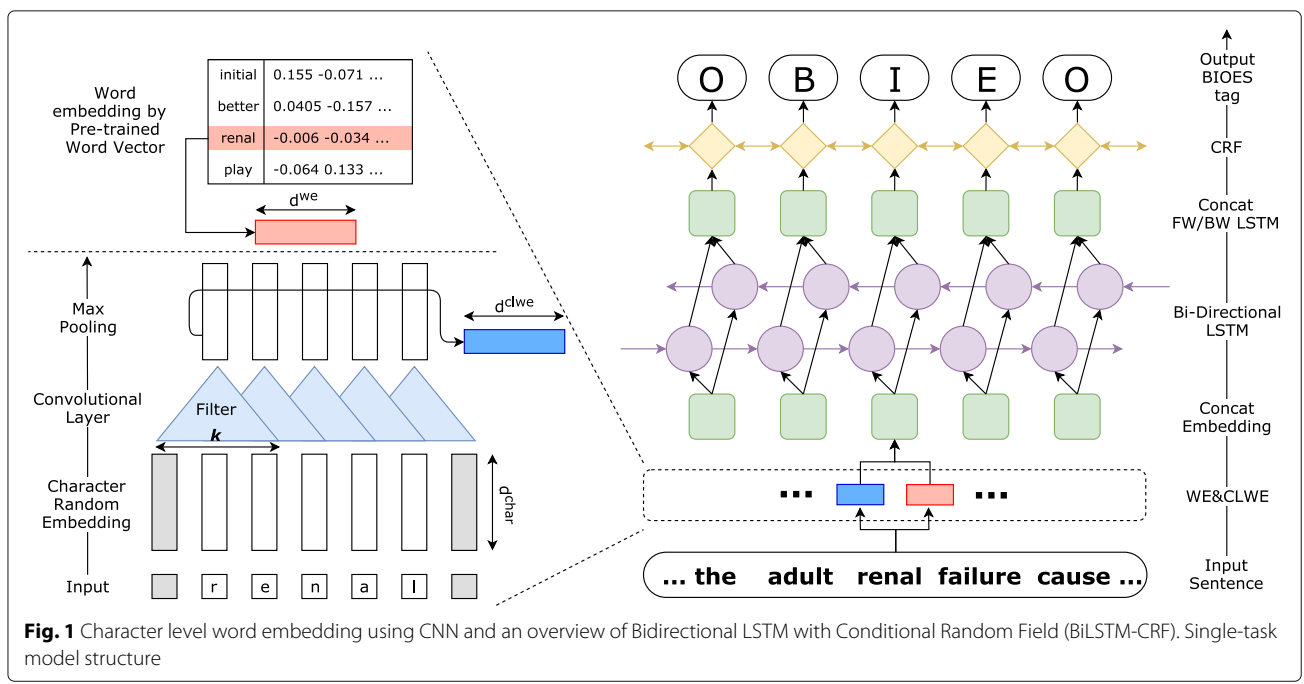

**Fig. 1** Character level word embedding using CNN and an overview of Bidirectional LSTM with Conditional Random Field (BiLSTM-CRF). Single-task model structure

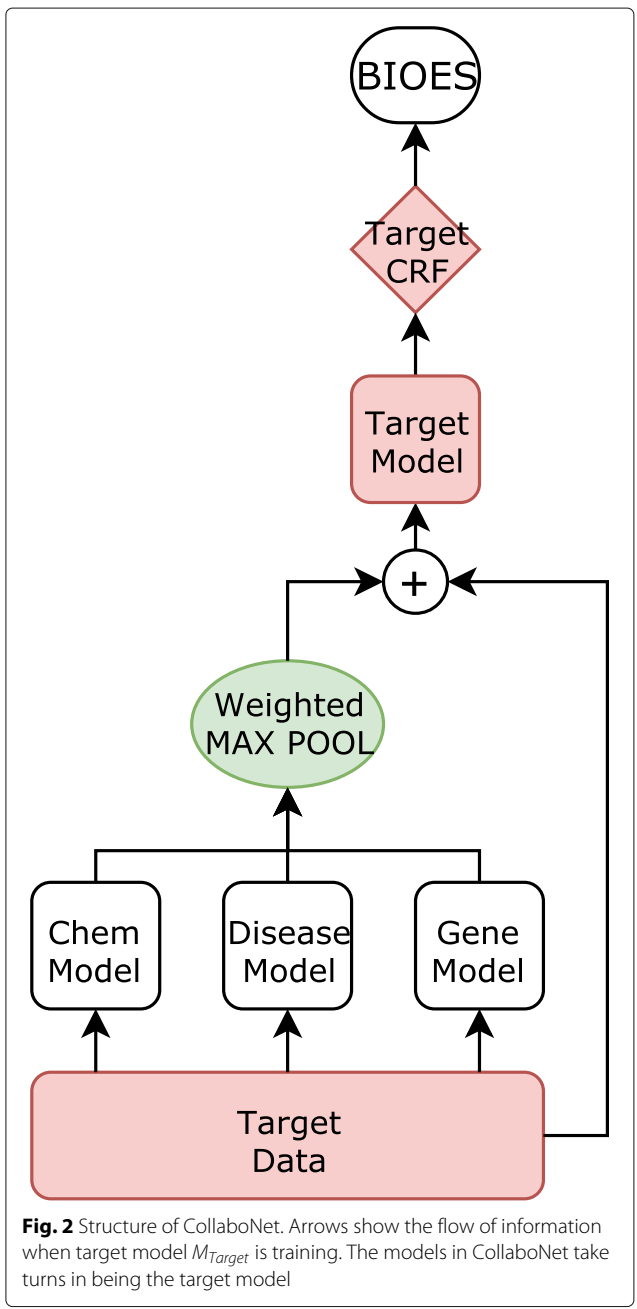

**Fig. 2** Structure of CollaboNet. Arrows show the flow of information when target model $M_{Target}$ is training. The models in CollaboNet take turns in being the target model

where $W_{char} \in \mathbb{R}^{d^{clwe} \times kd^{char}}$ and $b_{char} \in \mathbb{R}^{d^{clwe}}$ denote a trainable filter and bias, respectively. We obtain the element-wise maximum values, and the output is a character level word embedding denoted as $x_t^c \in \mathbb{R}^{d^{clwe}}$. We concatenate the character level word embedding with the word embedding trained on biomedical corpora as $\hat{x}_t = [x_t, x_t^c]$ to utilize both representations in our model.

**Long Short-Term Memory (LSTM)**
A Recurrent Neural Network (RNN) is a neural network that effectively handles variable-length inputs. RNNs have proven to be useful in various natural language processing tasks including language modeling, speech recognition and machine translation [28–30]. Long Short-Term Memory (LSTM) [31] is one of the most frequently used variants of recurrent neural networks. Our model uses the LSTM architecture from Graves et al. [29]. Given the outputs of an embedding layer $[\hat{x}_1, ..., \hat{x}_N]$, the hidden states of LSTM are calculated as follows:

$$i_t = \sigma \left( W_{xi}\hat{x}_t + W_{hi}h_{t-1} + b_i \right) \qquad (3)$$

$$f_t = \sigma \left( W_{xf}\hat{x}_t + W_{hf}h_{t-1} + b_f \right) \qquad (4)$$

$$c_t = f_t \odot c_{t-1} + i_t \odot \tanh \left( W_{xc}\hat{x}_t + W_{hc}h_{t-1} + b_c \right) \quad (5)$$

$$o_t = \sigma \left( W_{xo}\hat{x}_t + W_{ho}h_{t-1} + b_o \right) \qquad (6)$$

$$h_t = o_t \odot \tanh(c_t) \qquad (7)$$

where $\sigma$ and tanh denote a logistic sigmoid function and a hyperbolic tangent function, respectively, and $\odot$ is an element-wise product. We use a forward LSTM that extracts the representations of inputs in the forward direction, and we use a backward LSTM that represents the inputs in the backward direction.

We concatenate the two states coming from the forward LSTM and the backward LSTM to form the hidden states of the bi-directional LSTM (BiLSTM). BiLSTM, proposed by Schuster and Paliwal [32], was extensively used in various sequence encoding tasks. We obtain a set of hidden states $h_t^{bi} = \left[ h_t^f, h_t^b \right] \in \mathbb{R}^{2d^{lstm}}$ where $h_t^f$ and $h_t^b$ are hidden states of forward and backward LSTMs, respectively, at a time step $t$.

**Bidirectional LSTM with Conditional Random Field (BiLSTM-CRF)**
While BiLSTM handles long term dependency problems as well as backward dependency issues, modeling dependencies among adjacent output tags helps improve the performance of the sequence labeling models [25]. We applied a Conditional Random Field (CRF) to the output layer of the BiLSTM to capture these dependencies.

First, we compute the probability of each label given the sequence $S = [w_1, ..., w_N]$ as follows:

$$z_t = W_y h_t^{bi} + b_y \qquad (8)$$

$$\begin{aligned} p(y_t|w_1, ..., w_N; \Theta) &= \text{softmax}(z_t) \\ \text{softmax}(\mathbf{a}_j) &= \frac{\exp a_j}{\sum_k \exp a_k} \end{aligned} \qquad (9)$$

where $W_y \in \mathbb{R}^{5 \times 2d^{lstm}}$ and $b_y \in \mathbb{R}^5$ are parameters of the fully connected layer for BIOES tags, and the softmax(·) function computes the probability of each tag. Based on the probability $p$ and the CRF layer, our training objective to minimize is defined as follows:

$$L_{LSTM} = -\sum_{t=1}^{N} \log p(y_t|w_1, ..., w_N; \Theta) \quad (10)$$

$$L_{CRF} = -\sum_{t=1}^{T} \left( A_{y_{t-1}, y_t} + z_{t, y_t} \right) \quad (11)$$

$$Loss = L_{LSTM} + L_{CRF} \quad (12)$$

where $L_{LSTM}$ is the cross entropy loss for the label $y_t$, and $L_{CRF}$ is the negative sentence-level log likelihood. The score of a tag is the summation of the transition score $A_{y_{t-1}, y_t}$ and the emission score from our LSTM $z_{t, y_t}$ at time step $t$.

At test time, we use Viterbi decoding to find the most probable sequence given the outputs of the BiLSTM-CRF model.

### CollaboNet

CollaboNet, our novel NER model, is composed of multiple BiLSTM-CRF models (Fig. 2), and following the terminology of [25], we call each BiLSTM-CRF model a single-task model (STM). In CollaboNet, each STM is trained on a specific dataset and each STM is regarded as an expert on a particular entity type. These experts help each other since the knowledge of each expert is transferred to all the other experts. Training CollaboNet consists of phases and in each phase, except for the first preparation phase, only the target STM is trained on a single dataset for one epoch while the other STMs are not trained but only used to generate input for the target STM which is trained.

More formally, let us denote a set of datasets as $D$, and a single-task model as $M_k^n$, which is trained on the $k$-th dataset in phase $P^n$. In the preparation phase $(P^0)$ of CollaboNet, each STM is trained independently on a corresponding dataset until the performance of each model converges.

Note that an STM in the preparation phase $(M_k^0)$ is the same as a single BiLSTM-CRF model. In the preparation phase, we assume that each model $M_k^0$ has obtained the maximum amount of knowledge about the $k$-th dataset.

In the subsequent phases $P^n$, where $n \geq 1$, we select an STM $M_d^{n-1}$ which is an expert on the dataset $d$. We refer to the target STM $M_d^{n-1}$ as the *target model*, and the remaining STMs as the *collaborator models*. To train the target model $M_d^{n-1}$, we use inputs from the target dataset $d$ and BiLSTM outputs from collaborator models $M_k^{n-1}, \{k|k \neq d, k \in D\}$. We train each STM on its dataset for one epoch, and change the target STM $M_d^{n-1}$ as follows:

$$\hat{S}_d^n = \alpha_{k_1} M_{k_1}^{n-1}([S_d; \mathbf{0}]) \oslash \cdots \oslash \alpha_{k_m} M_{k_m}^{n-1}([S_d; \mathbf{0}]),$$
$$\{k_i | k_i \neq d, k_i \in S\} \quad (13)$$

$$\hat{p}(Y_d|S_d) = M_d^{n-1}\left([S_d; \hat{S}_d^n]\right) \quad (14)$$

where $[\cdot;\cdot]$ denotes concatenation and $\oslash$ denotes an aggregation operation such as max pooling or concatenation. We used weighted max pooling for the aggregation operation. $S_d$ is the input sequences of $d$-th dataset, and $M_d^{n-1}(\cdot)$ is output $h_t$, defined by Eq. 7. When aggregating the results of collaborator models, we multiply each of the results by a weight $\alpha_k$, which is a trainable parameter. The results are used to train the model $M_d^{n-1}$. Using the outputs obtained by Eq. 14, we train $M_d^{n-1}$ for one epoch, and it becomes $M_d^n$ in the next phase. The CRF layer is attached to the final output of $M_k^n$. Once we iterate all the target datasets $d \in D$, the next phase begins.

During the training phase $P^n$ for $d$, the target STM, which is composed of the BiLSTM layer and the CRF layer, and weights $\alpha_k$ $\{k|k \neq d, k \in D\}$ are trained. Parameters of the other STMs are not trained but the STMs generate only inferences on dataset $d$ in the training phase $P^n$. For example, when the disease dataset is the target dataset, the BiLSTM of the other STMs produces inferences about the other entity types for the disease dataset. More specifically, inferences about genes for the disease dataset $M_{gene}^{n-1}([S_{disease}; \mathbf{0}])$ which has rich information on gene entities, will benefit the disease STM.

## Experiments
### Datasets

We used 5 datasets (BC2GM [33], BC4CHEMD [34], BC5CDR [35–38], JNLPBA [22], NCBI [21]), all of which were collected by Crichton et al [24] (Table 1). Each of the 5 datasets were constructed from MEDLINE abstracts, and we used the BIOES notation format for named entity labels [26]. Each dataset focuses on one of the three biomedical entity types: disease, chemical, and gene/protein. We did not use cell-type entity tags from JNLPBA for the entity types.

All the datasets are comprised of pairs of input sentences and biomedical entity labels for the sentences. While the JNLPBA dataset has only training and test sets, the other four datasets contain training, development and test sets. For JNLPBA, we used part of its training set as its development set which is the same size as its test set. Also, we found that the JNLPBA dataset from Crichton et al. [24] contained sentences that were incorrectly split. So we preprocessed the original dataset by Kim et al. [22] with a more accurate sentence separation.

The BC5CDR dataset has the sub-datasets BC5CDR-chem, BC5CDR-disease and BC5CDR-both, and they contain chemical entity types, disease entity types, and both entity types, respectively. We reported the performance on BC5CDR-chem and BC5CDR-disease. We have a total of six datasets: BC2GM, BC4CHEMD, BC5CDR-chem, BC5CDR-disease, JNLPBA, and NCBI.

**Table 1** Descriptions of datasets

| Datasets | Entity type | # of sentence | # of annotations | Data Size |
| --- | --- | --- | --- | --- |
| NCBI-Disease (Dogan et al., 2014) | Disease | 7639 | 6881 | 793 abstracts |
| JNLPBA (Kim et al., 2004) | Gene/Proteins | 22,562 | 35,336 | 2404 abstracts |
| BC5CDR (Li et al., 2016) | Chemicals | 14,228 | 15,935 | 1500 articles |
| BC5CDR (Li et al., 2016) | Diseases | 14,228 | 12,852 | 1500 articles |
| BC4CHEMD (Krallinger et al., 2015a) | Chemicals | 86,679 | 84,310 | 10,000 abstracts |
| BC2GM (Akhondi et al., 2014) | Gene/Proteins | 20,510 | 24,583 | 20,000 sentences |

**Metric**

For the evaluation of the named entity recognition task, true positives are counted from exact matches between predicted entity spans and ground truth spans based on the BIOES notation.

We also designed and applied a simple post-processing step that corrects invalid BIOES sequences. This simple step improved precision by about 0.1 to 0.5%, and thus boosted the F1 score by about 0.04 to 0.3%.

Precision, recall and F1 scores were used to evaluate the models.

- M = total number of predicted entities in the sequence.
- N = total number of ground truth entities in the sequence.
- C = total number of correct entities.

$$Precision = P = \frac{C}{M}, Recall = R = \frac{C}{N}, \\ F_1 score = \frac{2PR}{P+R} \quad (15)$$

**Settings and hyperparameters**

We used the 200 dimensional word embedding (WE) by Pyysalo et al. [20] which was trained on PubMed, PubMed Central (PMC) and Wikipedia text, and it contains about 5 million words. Word2vec [39] was used to train the word embedding. For character level word embedding (CLWE), we used window sizes of 3, 5, and 7.

We used AdaGrad optimizer [40] with an initial learning rate of 0.01 which was exponentially decayed for each epoch by 0.95. The dimension of the character embedding ($d^{char}$) was 30 and dimension of the character level word embedding ($d^{clwe}$) was 200*3. We used 300 hidden units for both forward and backward LSTMs. Weights for aggregating the results of collaborator models were uniformly initialized with 1. We applied dropout [41] to two parts of CollaboNet: output of CLWE (0.5) and output of BiLSTM (0.3). The mini-batch size for our experiment was 10.

Most of our hyperparameter settings are similar to those of Wang et al. [25]. Only a few settings such as the dropout rates were different from the hyperparameters of Wang. We tuned these hyperparameters using validation sets.

The preparation phase $P^0$ for 6 datasets takes approximately 900 min, which is the same amount of time it takes to train 6 single-task models. The rest of the phases $P^n, n \geq 1$ require 3000 min for complete training. If we exclude BC4CHEMD, the largest dataset, then the training time for $P^n$ is reduced to 1500 min, which is half the time required for the remainder phases. Experiments were conducted on a 10-core CPU (Intel Xeon E5-260 v4 CPU 2.2 GHz) with one graphics processing unit (NVIDIA Titan Xp). Our code is written in TensorFlow 1.7 (GPU enabled version) for Python 2.7.

**Results**

The experimental results of the baseline models and CollaboNet are provided in Tables 2 and 3, respectively. Table 2 shows the results of the single-task models (STMs) where Table 3 shows the comparison between the existing state-of-the-art multi-task learning model (MTM) and our CollaboNet.

Since Wang et al. [25] used BC5CDR-both for their experiments, we reran their models on BC5CDR-chem and BC5CDR-disease for a fair comparison with other models. The rerun scores are denoted with asterisks. We conducted 10 experiments with 10 different random initializations on our STM. We take arithmetic mean over the 6 datasets to compare the overall performance of each model.

**Performance of single-task models**

Table 2 shows the results of the STMs of Habibi et al. [18] and Wang et al. [25] (baseline STMs), and our STM on the 6 datasets. While the baseline STMs applied BiLSTM for the Character Level Word Embedding (CLWE) layer [18, 25], our STM used Convolution Neural Network (CNN) for the CLWE layer.

**Table 2** Performances of single-task models

| Model | Habibi et al. (2017) STM | | | Wang et al. (2018) STM | | | Our STM | | |
|---|---|---|---|---|---|---|---|---|---|
| Dataset | Precision | Recall | F1 Score | Precision | Recall | F1 Score | Precision | Recall | F1 Score |
| NCBI-disease | 85.31 | 83.58 | 84.44 | 84.95 | 82.92 | 83.92 | 83.95 | 85.45 | **84.69** (±0.54) |
| JNLPBA | 74.83 | 79.82 | 77.25 | 69.60 | 74.95 | 72.17 | 72.51 | 82.98 | **77.39** (±0.24) |
| BC5CDR-chem | 92.57 | 88.77 | 90.63 | *93.05 | *86.87 | *89.85 | 94.02 | 91.50 | **92.74** (±0.47) |
| BC5CDR-disease | 84.19 | 82.79 | **83.49** | *84.09 | *81.32 | *82.68 | 82.98 | 82.25 | 82.61 (±0.25) |
| BC4CHEMD | 87.83 | 85.45 | 86.62 | 90.53 | 87.04 | **88.75** | 90.50 | 85.96 | 88.19 (±0.23) |
| BC2GM | 77.50 | 78.13 | 77.82 | 81.11 | 78.91 | **80.00** | 79.70 | 77.47 | 78.56 (±0.38) |
| Macro Average | 83.71 | 83.09 | 83.38 | 83.89 | 82.00 | 82.90 | **83.94** | **84.27** | **84.03** |

Our STM achieved the best performance on 3 datasets among 6. Scores in the asterisked (*) cells are obtained in the experiments that we conducted; these scores are not reported in the original papers. The best scores from these experiments are in bold

On average, our STM outperforms the baseline STMs in terms of precision, recall and F1 score. Although, Sahu and Anand [17] tried to improve the performance of NER models with CNN based CLWE layer, they have failed to do so. In our experiments, however, our STM outperforms other baseline STMs, demonstrating the effectiveness of STM with CNN based CLWE layer.

**Performance of CollaboNet**

Comparing Tables 2 and 3, CollaboNet achieves higher precision and F1 score than most STM models on all datasets. On average, CollaboNet has improved both precision and recall. CollaboNet also outperforms the multi-task model (MTM) from Wang et al. [25] on 4 out of 6 datasets (Table 3). While multi-task learning has improved performance in previous studies [25], using CollaboNet, which consists of expert models trained for each entity type, could further improve biomedical named entity recognition performance.

**Discussion**

Compared to baseline models, CollaboNet achieves higher performance on macro average (Tables 2 and 3). The increase in precision is supportive when considering the practical use of the bioNER systems. In a number of biomedical text mining systems, important information tends to be repeated in a large size text corpus. Therefore, missing a few entities may not hinder the performance of an entire system, as this can be compensated elsewhere. However, incorrect information and the propagation of errors can effect the entire system.

In Table 4, we report the error types of our STM and CollaboNet. We define *bio-entity error* as recognizing different types of biomedical entities as target entity types. For instance, recognizing '*VHL*' as a gene when it was used as a disease in a sentence is a bio-entity error. Note that a bio-entity error could occur when an entity is a polysemous word (e.g. VHL), or comprised of multiple words (e.g. BRCA1 deficient), and thus correcting bio-entity errors requires contextual information or supervision of other entity type models. The error analysis was conducted on 4334 errors of our STM and 3966 errors of CollaboNet on 5 datasets (BC2GM, BC5CDR-chem, BC5CDR-disease, JNLPBA, NCBI). Error analysis was conducted on models which showed best performance in our experiments.

The error analysis of our STM, which is a single BiLSTM-CRF model, shows that the majority of errors

**Table 3** Performance of CollaboNet and the Multi-Task Model by Wang et al. [25]

| Model | Wang et al. (2018) MTM | | | CollaboNet | | |
|---|---|---|---|---|---|---|
| Dataset | Precision | Recall | F1 Score | Precision | Recall | F1 Score |
| NCBI-disease | 85.86 | 86.42 | 86.14 | 85.48 | 87.27 | **86.36** (±0.54) |
| JNLPBA | 70.91 | 76.34 | 73.52 | 74.43 | 83.22 | **78.58** |
| BC5CDR-chem | *93.09 | *89.56 | *91.29 | 94.26 | 92.38 | **93.31** |
| BC5CDR-disease | *83.73 | *82.93 | *83.33 | 85.61 | 82.61 | **84.08** |
| BC4CHEMD | 91.30 | 87.53 | **89.37** | 90.78 | 87.01 | 88.85 |
| BC2GM | 82.10 | 79.42 | **80.74** | 80.49 | 78.99 | 79.73 |
| Macro Average | 84.50 | 83.70 | 84.07 | **85.18** | **85.25** | **85.15** |

Scores in the asterisked (*) cells are obtained in the experiments that we conducted; these scores are not reported in the original papers. The best scores from these experiments are in bold

**Table 4** The number of bio-entity type errors, the total number of errors, and the ratio of bio-entity errors to the total numbers of errors for each model prediction

| Dataset | Our STM | | | CollaboNet | | | Difference |
|---|---|---|---|---|---|---|---|
| | Bio Entity | Total | Ratio of Bio Entity | Bio Entity | Total | Ratio of Bio Entity | |
| NCBI-disease | 54 | 167 | 32.3% | 38 | 131 | 29.0% | -3.3% |
| JNLPBA | 749 | 1520 | 49.3% | 227 | 1437 | 15.8% | **-33.5%** |
| BC5CDR-chem | 142 | 503 | 28.2% | 122 | 505 | 24.2% | -4.1% |
| BC5CDR-disease | 199 | 867 | 23.0% | 131 | 728 | 18.0% | -5.0% |
| BC2GM | 189 | 1277 | 14.8% | 218 | 1165 | 18.7% | 3.9% |

Negative values at the difference tab indicate that CollaboNet reduced the number of false positives, especially false biomedical entities are classified as bio-entity errors which comprise up to 49.3% of the total errors in JNLPBA. According to the error analysis of our STM model, bio-entity errors constitute 1333 errors out of 4334 errors, comprising 30.8% of all the errors. Although bio-entity error was not the most common error type, the importance of bio-entity error is much greater that of other errors such as span error which was the most common error type, constituting 38% of incorrect errors. While most span errors can be easily fixed by non-experts, bio-entity errors are difficult to detect and fix, even for biomedical researchers. Also, for biomedical text mining tasks such as drug-drug interaction (DDI) extraction, span errors of an NER system have a minor effect on DDI results but bio-entity errors could lead to completely different results.

The performance improvement of CollaboNet over STM may not seem significant when considering the increased complexity of CollaboNet's structure. We found by error analysis that CollaboNet had an increased number of span errors. As our metric is based on the exact match evaluation, consistent annotation of the ground truth dataset is important for reducing span errors which are caused by modifiers. For instance, in the phrase "acute adult renal failure," "adult renal failure" may be labeled as an entity in some datasets. In this case, predicting "acute adult renal failure" or "renal failure" as an entity will be counted as a false negative and a false positive. On the other hand, some other datasets may include the modifier "acute" in an entity, considering "acute adult renal failure" as the only true prediction. Therefore, unlike STM, CollaboNet uses various datasets that have been annotated differently. Even though CollaboNet outperforms STM, its results may be lower due to this inconsistency in annotation.

In CollaboNet, each expert model is trained on a single entity type dataset, and their training inputs are a concatenation of word embeddings and outputs of the other expert models. We expect that the other expert models will transfer knowledge on their respective entity to the target model, and thus improve the bio-entity type error problem by collaboration. As Table 4 shows, CollaboNet performs better than our STM in detecting polysemy and other entity types. Among 3966 errors from CollaboNet, 736 errors are bio-entity errors, comprising 18.6% of all the errors.

### Case study

We sampled the predictions of CollaboNet and those of our STM (single-task model) to further understand the strengths of CollaboNet in Table 5.

The first example from chemical dataset in Table 5 shows our expected result from CollaboNet. Our STM annotates *antilymphocyte globulin* as a chemical entity. However, it is clear that the entity is not a chemical but a type of globulin which is a protein. The second example sentence from the chemical dataset is about an *ACE / ARB* entity. Again, our STM misidentifies the entity as a chemical entity. On the other hand, in CollaboNet, the target model (chemical model) obtains knowledge from one of the collaborator models (the gene/protein model) to avoid mistakenly recognizing the entity as a chemical entity. As *globulin* or *ACE* entities appear in the gene/protein dataset, the chemical model obtains information from the gene/protein model.

In the disease dataset, the first example shows a multi-word entity in parentheses. As a gene model can pass syntactic and semantic information about a word *e.g., mutated* and its surrounding words to a disease model, CollaboNet can abstain from predicting *A-T, mutated* as the disease entity, which our STM model failed to do. The second example in the disease dataset is on *cardiac troponin T*. Since *cardiac + noun* in biomedical text can be easily considered as a disease name, our STM misidentified this word as a disease entity. However, with the help of a gene model, CollaboNet did not mark it as a disease entity.

The gene/protein entity type further demonstrates the effectiveness of CollaboNet in reducing bio-entity type errors. Two example sentences contain abbreviations, which are one of the distinct characteristics

**Table 5** Case study

| | Chemical dataset | |
|---|---|---|
| Our STM | No prophylaxis with <u>antilymphocyte globulin</u> was used | - globulin : Protein |
| CollaboNet | No prophylaxis with antilymphocyte globulin was used | |
| Ground Truth | No prophylaxis with antilymphocyte globulin was used | |
| Our STM | elderly patients using <u>ACE / ARB</u> in combination with <u>potassium</u> | ACE : Gene/Protein |
| CollaboNet | elderly patients using ACE / ARB in combination with <u>potassium</u> | |
| Ground Truth | elderly patients using ACE / ARB in combination with <u>potassium</u> | |
| | Disease Dataset | |
| Our STM | The ATM (<u>A-T, mutated</u>) gene on human chromosome 11q22. | A-T, mutated : Gene |
| CollaboNet | The ATM (A-T, mutated) gene on human chromosome 11q22. | |
| Ground Truth | The ATM (A-T, mutated) gene on human chromosome 11q22. | |
| Our STM | to bind to the human <u>cardiac troponin T</u> (cTNT) pre-messenger RNA | cTNT : Gene/Protein |
| CollaboNet | to bind to the human cardiac troponin T (cTNT) pre-messenger RNA | |
| Ground Truth | to bind to the human cardiac troponin T (cTNT) pre-messenger RNA | |
| | Gene / Protein Dataset | |
| Our STM | which is inhibited by the <u>cytotoxin leptomycin B (LMB)</u>, and also by its interaction | LMB : Chemical, Drug |
| CollaboNet | which is inhibited by the cytotoxin leptomycin B (LMB), and also by its interaction | |
| Ground Truth | which is inhibited by the cytotoxin leptomycin B (LMB), and also by its interaction | |
| Our STM | Classic Hodgkin disease (<u>cHD</u>) is derived from B cells with high loads of mutations. | cHD : Disease |
| CollaboNet | Classic Hodgkin disease (cHD) is derived from B cells with high loads of mutations | |
| Ground Truth | Classic Hodgkin disease (cHD) is derived from B cells with high loads of mutations | |

This table contains sentences that were incorrectly predicted by of our STM but were correctly predicted by CollaboNet. The predicted labels or the ground truth labels are underlined of gene entities. *LMB* and *cHD* are incorrectly predicted as gene/protein entities by our STM, since lots of gene/protein entities are abbreviations. However, the target model (gene/protein model) in CollaboNet can obtain information on *leptomycin* and *disease* from the chemical and disease models, respectively. With the help of information from collaborator models, CollaboNet can effectively increase the precision of other entity type models.

In addition, we found some labels in the ground truth set, which we believe are incorrect. Tsai et al. [15] also reported that the inconsistent annotations in the JNLPBA corpus limit the NER system. We report our findings in Table 6.

**Table 6** Case study

| | Gene / Protein Dataset |
|---|---|
| CollaboNet | Troglitazone, a <u>PPARgamma ligand</u>, inhibits <u>osteopontin gene</u> expression in THP-1 cells. |
| Ground Truth | Troglitazone, a PPARgamma ligand, inhibits osteopontin gene expression in THP-1 cells |
| CollaboNet | The <u>translesion DNA polymerase zeta</u> plays a major role in <u>Ig</u> and <u>bcl-6</u> somatic hypermutation. |
| Ground Truth | The <u>translesion DNA polymerase zeta</u> plays a major role in Ig and bcl-6 somatic hypermutation. |
| | Chemical Dataset |
| CollaboNet | recently identified Delta22-isomer of <u>beta-muricholate</u> contribute for 5.4% |
| Ground Truth | recently identified Delta22-isomer of beta-muricholate contribute for 5.4% |
| CollaboNet | <u>Hexabrix and polyvidone</u> are considered the best contrast media for hysterosalpingography. |
| Ground Truth | <u>Hexabrix and polyvidone</u> are considered the best <u>contrast media</u> for hysterosalpingography. |

This table shows the questionable answers from the ground truth datasets. Our model achieves better performance in detecting entities in these example sentences. The predicted labels or the ground truth labels are underlined

In the first row of Table 6, the gene/protein entity *osteopontin* was not marked in the ground truth labels, whereas our network correctly predicted it as a gene entity. The second row also displays questionable results of the ground truth labels. Although *lg* and *bcl-6*, which are abbreviations of *Immunoglobulin* and *B-cell lymphoma 6*, where not labeled in the ground truth labels, our model detected them as a gene / protein entity. The example sentences of gene/protein annotations in Table 6 were reviewed by several domain experts and medical doctors. As shown in the third row, *beta-muricholate* is a chemical entity but it was not annotated in the ground truth labels. However, the last row shows another type of annotation error. *Contrast media* is a general term for a medium used in medical imaging and since is not a proper noun, it is not a named entity.

These examples shows the presence of incorrect ground truth labels, which can harm the performance of bioNER models. However, we believe that these missed or misidentified ground truth labels can be corrected by our system.

### Future works

For future work, we plan to cover more target entity types and use more datasets. For example, CRAFT [42], LINNAEUS [43] and Variome [44] are manually annotated datasets and are valuable resources that can be used for expanding our model. Second, we plan to apply CollaboNet to downstream biomedical text mining systems. For example, entity search engines such as BEST [10] could be improved by using more accurate NER models.

## Conclusion

In this paper, we introduced CollaboNet, which consists of multiple BiLSTM-CRF models, for biomedical named entity recognition. While existing models were only able to handle datasets with a single entity type, CollaboNet leverages multiple datasets and achieves the highest F1 scores. Unlike recently proposed multi-task models, CollaboNet is built upon multiple single-task NER models (STMs) that send information to each other for more accurate predictions. In addition to the performance improvement over multi-task models, CollaboNet differentiates between biomedical entities that are polysemous or have similar orthographic features. As a result, our model achieved state-of-the-art performance on four bioNER datasets in terms of F1 score, precision and recall. Although our model requires a large amount of memory and time, which existing multi-task models require as well, the simple structure of CollaboNet allows researchers to build another expert model for different entity types in CollaboNet. As CollaboNet obtains higher precision than other models, we plan to apply CollaboNet in a biomedical text mining system.

**Abbreviations**
BiLSTM: Bidirectional long short-term memory; BioNER: Biomedical named entity recognition; CE: Character embedding; CLWE: Character level word embedding; CNN: convolution neural network; CRF: Conditional random field; DDI: Drug-Drug Interaction; LSTM: long short-term memory; MTL: Multi-task learning; MTM: Multi-task model; NER: Named entity recognition; NLP: Natural language processing; PMC: PubMed Central; STM: Single-task model; RNN: Recurrent neural network; WE: Word embedding

**Acknowledgements**
We are sincerely grateful to Inah Chang for conducting manual error counting. We appreciate Susan Kim for editing the manuscript.

**Funding**
The design of the study and collection, analysis, and interpretation of data were funded by the National Research Foundation of Korea (NRF-2017M3C4A7065887, 2016M3A9A7916996) and National IT Industry Promotion Agency grant funded by the Ministry of Science and ICT and Ministry of Health and Welfare (NO. C1202-18-1001, Development Project of The Precision Medicine Hospital Information System (P-HIS)). Publication costs were funded by the National Research Foundation of Korea (NRF-2016M3A9A7916996).

**Availability of data and materials**
The source code of CollaboNet and the datasets are available at https://github.com/wonjininfo/CollaboNet.

**About this supplement**
This article has been published as part of *BMC Bioinformatics Volume 20 Supplement 10, 2019: Proceedings of the 12th International Workshop on Data and Text Mining in Biomedical Informatics (DTMBIO 2018)*. The full contents of the supplement are available online at https://bmcbioinformatics.biomedcentral.com/articles/supplements/volume-20-supplement-10.

**Authors' contributions**
WY, CHS, JL and JK conceived the idea. WY and JL designed the model. WY and CHS developed CollaboNet. CHS experimented and collected analysis examples and results. WY, JL and JK wrote the manuscript. JK, as the supervisor of WY, CHS and JL, provided guidance on the experiment. All authors read and approved the final manuscript.

**Ethics approval and consent to participate**
Not applicable.

**Consent for publication**
Not applicable.

**Competing interests**
The authors declare that they have no competing interests.

**Publisher's Note**
Springer Nature remains neutral with regard to jurisdictional claims in published maps and institutional affiliations.

Published: 29 May 2019